%% file: iclr2024_arxiv.tex
\documentclass{article} %
\usepackage{iclr2024_conference,times}

\input{math_commands.tex}

\usepackage{hyperref}
\usepackage{listings}
\usepackage{caption}
\usepackage{url}
\usepackage{wrapfig}
\usepackage{graphicx} 
\usepackage{subcaption}
\usepackage{bbm}
\usepackage{amsmath}
\usepackage{amsfonts}
\usepackage{amsthm}
\usepackage{booktabs}

\newcommand{\sys}[1]{\color{red!70!black}\textbf{#1}}
\newcommand{\usr}[1]{\color{blue!70!black}\textbf{#1}}
\newcommand{\astnt}[1]{\color{green!70!black}\textbf{#1}}
\newtheorem{theorem}{Theorem}

\title{Language Models Can Reduce Asymmetry in Information Markets}

\iclrfinalcopy
\author{Nasim Rahaman$^{*, 1, 2}$ $\quad$ Martin Weiss$^{*, 1, 3}$ $\quad$ Manuel W\"urthrich$^{4}$ \AND Yoshua Bengio$^{1, 5}$ $\quad$ Li Erran Li$^{\dagger 6}$ $\quad$ Chris Pal$^{\dagger, 1, 3}$ $\quad$ Bernhard Sch\"olkopf$^{\dagger, 2}$}

\begin{document}

\maketitle

\begin{abstract}

This work addresses the buyer's inspection paradox for information markets. The paradox is that buyers need to access information to determine its value, while sellers need to limit access to prevent theft. To study this, we introduce an open-source simulated digital marketplace where intelligent agents, powered by language models, buy and sell information on behalf of external participants. The central mechanism enabling this marketplace is the agents' dual capabilities: they not only have the capacity to assess the quality of privileged information but also come equipped with the ability to forget. This ability to induce amnesia allows vendors to grant temporary access to proprietary information, significantly reducing the risk of unauthorized retention while enabling agents to accurately gauge the information's relevance to specific queries or tasks. To perform well, agents must make rational decisions, strategically explore the marketplace through generated sub-queries, and synthesize answers from purchased information. Concretely, our experiments (a) uncover biases in language models leading to irrational behavior and evaluate techniques to mitigate these biases, (b) investigate how price affects demand in the context of informational goods, and (c) show that inspection and higher budgets both lead to higher quality outcomes.
\footnotetext{$^1$ Mila, Quebec AI Institute$\quad$$^2$ Max Planck Institute for Intelligent Systems, T\"ubingen$\quad$$^3$ Polytechnique Montr\'eal$\quad$$^4$ Harvard SEAS$\quad$$^5$ Universit\'e de Montr\'eal$\quad$$^6$ Amazon Web Services$\quad$$^{*\,\dagger}$ Equal Contribution}

\end{abstract}

\section{Introduction}
\vspace{-4pt}
Information economics is the study of how systems of information affect economic decisions and outcomes. A core challenge in designing mechanisms for information markets stems from the fact that information is often expensive to produce but cheap to reproduce \citep{samuelson2009economics}. To address this, information producers across industries deploy barriers to entry via paywalls, tiered subscription models, and advertisements in an attempt to balance monetization and reach. However, such mechanisms inherently impede accessibility and discovery of information. Viewed through the lens of Information Foraging Theory \citep{card1999}, these mechanisms obstruct the ``information trail'' — the cues guiding users to valuable sources of information — thus compromising the information-seeking process.

In the light of rising obstructions to information discovery, it is unsurprising that users are flocking towards large language model (LLM) powered tools. These LLMs help users navigate the information trail, providing them with both high-level overviews and low-level details as needed, while tailoring responses based on their past inquiries and level of expertise. However, LLMs are also problematic. They are trained on massive datasets compiled from the internet and elsewhere by automated scrapers \citep{openai2023gpt4, touvron2023llama, pile2021}, and they can internalize and reproduce monetized information. This practice has understandably raised concerns about the unauthorized dissemination of proprietary and copyrighted content \citep{alter2023openainyt}. In response, content providers are employing additional legal and technical barriers, further exacerbating the content discovery problem for consumers of information.

In this work, we focus on enabling information flow by addressing a key challenge in information markets -- the buyers' inspection paradox \citep{arrow1972}. This paradox requires that buyers need access to information to assess its value, but sellers need to limit access to information to prevent expropriation. The paradox leads to information asymmetry between sellers (who know more about the quality of their information) and buyers (who know less), exacerbating adverse selection mechanisms. This context closely mirrors the dilemma in used car markets, outlined in the \textit{The Market for ``Lemons": Quality Uncertainty and the Market Mechanism}~\citep{akerlof1970}. Sellers, holding superior information about their goods, unintentionally force buyers into a defensive position, paying a higher price to often receive lower-quality goods. This creates a cycle that further devalues the goods, nudging the market towards a collapse.

\textbf{A central argument} of this paper asserts that artificial agents, powered by language models, can contribute to mitigating the pervasive issue of information asymmetry in information markets. These agents come with dual capabilities: a capacity to evaluate the quality of privileged information and the ability to forget. By granting these agents temporary access to proprietary information, vendors significantly reduce the risk of expropriation while allowing the agents to assess the information's relevance to a specific query or task. If the accessed information is judged to be non-essential, duplicative, or more expensive than substitutable information from a cheaper source,  the agent can choose to discard it without incurring acquisition costs, ensuring no unauthorized retention. Conversely, if the information proves valuable, the agent can make an informed decision to buy. Purchasing information not only enhances the agent's ability to synthesize an answer, but can also provide a basis for sub-queries, following the trail to a better answer.

Concretely, we pose three research questions:
\begin{enumerate}
    \item Is it possible to establish a functional digital marketplace for information where language model based agents preview, value, and purchase information on behalf of their principals?
    \item Does this marketplace enable buyers to more reliably identify and value information?
    \item How do language model based agents behave as economic actors? What biases are they subject to as they value and procure information? 
\end{enumerate}

\textbf{The primary technical contribution} of this work is an open-source \footnote{\url{https://github.com/tn-learn/info-bazaar}} text-based environment where we evaluate agents' capacity to perform tasks related to operating as economic actors in an information market. The environment emulates a marketplace populated by buyers and vendors, implemented by LLM agents. Principals have questions to answer and a budget of market credits, and buyer agents are appointed on their behalf. Concurrently, vendor agents represent content providers who possess a repository of documents they have the rights to and are willing to sell access to. This simulated marketplace has been constructed to allow buyer agents the flexibility to peruse information without binding commitment to purchase. Furthermore, it can permit a buyer agent to purchase information in multiple rounds before answering their principal. To support the environment, we have collected a dataset consisting of 725 research papers concerning LLMs sourced from ArXiv. To evaluate the quality of answers generated by buyer agents within different scenarios, synthetic questions were formulated by an LLM, undergoing filtering and deduplication. 

\begin{figure}
    \centering
    \includegraphics[width=\textwidth]{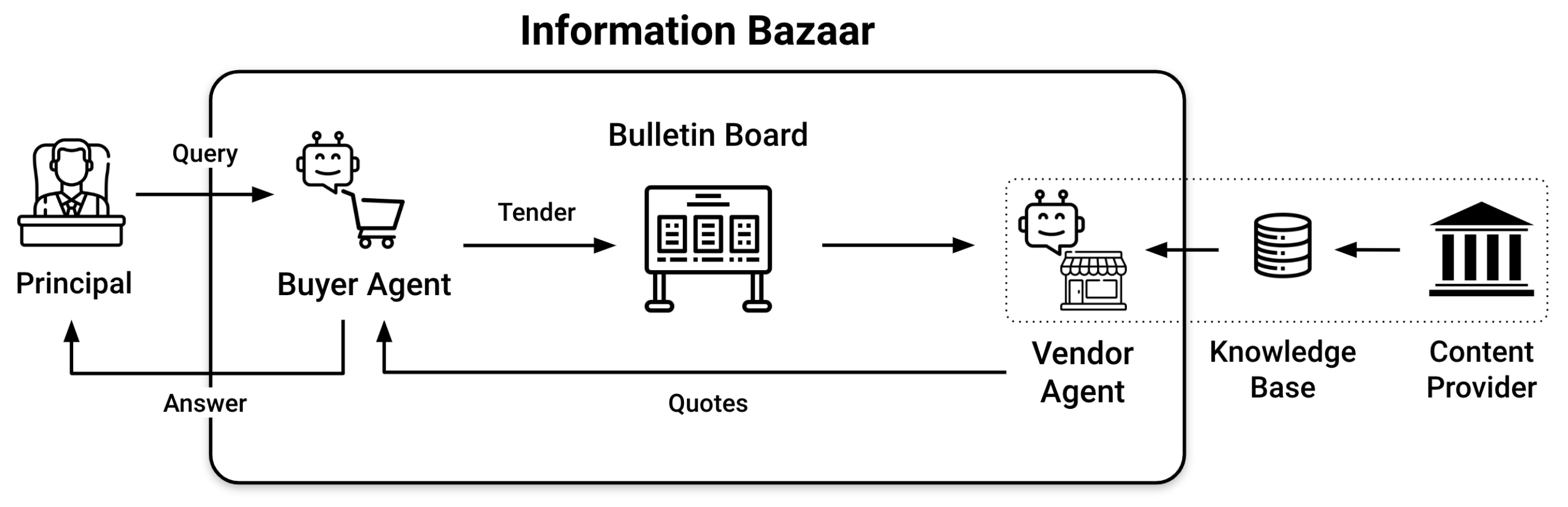}
    \caption{
    \looseness=-1
    \textbf{The Information Bazaar} is a simulated marketplace for information. Principals authorize buyer agents to answer a query (a question and budget). The process starts with buyer agents posting tenders (requests for specific information) on a Bulletin Board. Vendor agents, holding information from various external sources, assess these tenders and may respond with quotes (i.e., their priced information offers). Buyer agents then evaluate these quotes. If they decide not to purchase specific information, then they immediately forget it, ensuring only purchased information is retained for further use. The cycle of posting tenders, receiving, and assessing quotes continues, with buyer agents optionally forming sub-queries based on purchased information to seek deep insights. The agents work within this framework until they compile satisfactory answers, exhaust their budget, or reach a pre-set tree-size limit in the Bazaar. The final step involves the buyer agents synthesizing a comprehensive answer for their principals, using only the information they have purchased.
    \textbf{tl;dr} In the Information Bazaar, agents continuously navigate information exchange, only retaining and utilizing purchased information to derive comprehensive and satisfactory answers within given constraints.
    }
    \label{fig:fig-1}
    \vspace{-10pt}
\end{figure}

\section{Related Work}
\vspace{-4pt}
\textbf{Information Economics}. The challenge of valuing information has long occupied economists, particularly in the context of information asymmetry and market inefficiencies. Seminal work by \citet{akerlof1970} demonstrated how asymmetric information can disrupt markets, while \citet{stigler1961} focused on market failures due to the obstructed information flows. Central to our work is Arrow's concept of the buyer's inspection paradox \citep{arrow1972, }, which explores the dilemma of valuing information that one cannot fully inspect before purchase. Our Information Bazaar addresses this paradox with agents that reliably forget unpurchased information.

\textbf{Information Foraging and Retrieval}. Information Foraging Theory (IFT) serves as a metaphorical framework likening information consumption to animal foraging. Works by \citet{pirolli2007} and \citet{card1999} have employed IFT to understand cues and decision-making in information-seeking. LLMs have demonstrated remarkable capabilities in information extraction and retrieval \citep{radford2019, retro2021, openai2023gpt4, bai2022training}, which our work leverages to appraise information. Systems like Baleen \citep{Vaughan2021} and the approach proposed by \citet{yogatama2021} introduce methods for multi-document retrieval and complex query handling, which share similarities with our agentic retrieval approach.

\textbf{Agent Models and Marketplace Simulation}. Work on simulating digital marketplaces and multi-agent systems is directly related to our work, including the consideration of artificial economies for autonomous agents by \citet{wellman2001} and principles of multi-agent systems by \citet{woolridge2001}. Similar to \citet{zheng2020ai}, we engage in testing economic policies in simulated environments, and akin to \citet{horton2023homosilicus}, we utilize LLMs as simulated economic agents. Despite these similarities, our work distinctly assumes that the Information Bazaar acts as a market regulator, controlling the behavior of buyer agents to prevent information expropriation. The use of LLMs for rational decision-making has also been explored by frameworks such as AgentBench \citep{liu2023agentbench} and REACT \citep{yao2023react}, which explore complex, multi-turn interactions, inspiring the technical design of the Information Bazaar.

\section{The Information Bazaar}
\vspace{-4pt}
In this section, we introduce a text-based environment, termed the Information Bazaar\footnote{In a physical bazaar, buyers have the ability to inspect goods without a binding commitment to purchase.}. This environment is a synchronous simulation of an information marketplace populated by buyer and vendor agents (see Section~\ref{sec:agents} for details). The marketplace infrastructure includes provisions for buyer agents to place tenders, to which vendor agents can respond with quotes, as outlined in Section~\ref{sec:tenders-and-quotes}. Notably, buyer agents possess the capability to pose follow-up questions, enabling them to delve deeper into the information landscape, as discussed in Section~\ref{sec:info-trail}. We implemented the Information Bazaar in Python, utilizing the mesa library \citep{kazil2020mesa}, a library for agent-based modeling. The particular instantiation of the bazaar analyzed uses a dataset comprising 725 research papers on LLMs sourced from Arxiv, elaborated on in Section~\ref{sec:impl-deets}. For a mathematical analysis detailing the theoretical impacts of inspection on expected utility within the bazaar, refer to Appendix~\ref{app:math}.

\subsection{Principals and Agents} \label{sec:agents}
\vspace{-4pt}
We primarily classify agents into two categories: buyers and vendors. Buyer agents have a set question and budget, aiming to find the best answer for the lowest cost. Vendor agents sell documents for their principals, aiming to earn market credits. Each of these documents carries its own price tag.  We allow for two or more content providers to possess and independently price the same piece of information. As a simplification, we do not simulate affordances for vendors to modify the prices in response to demand. We leave analysis of pricing strategies to future work. 

The roles and objectives of these agents are clearly delineated. A buyer agent's primary mission is to navigate the bazaar to obtain the most accurate and complete answer to its principal's query without overspending its allocated market credits. To achieve this, the agent must engage in transactions with vendor agents to access the necessary information. On the other hand, the vendor agent's role is to sell the information held by their content provider. By doing so, they aim to accumulate market credits, which accrue to the benefit of their content provider.

\subsection{Tenders and Quotes} \label{sec:tenders-and-quotes}
\vspace{-4pt}
The process of information exchange commences when a buyer agent posts a tender to the bulletin board. Each tender consists of a query and a maximum budget that the agent is authorized to spend. Upon seeing these tenders, vendor agents engage in what we term the vendor-side retrieval process, wherein they sift through their principal's repository of documents\footnote{To simplify the exposition, ``documents'' and ``passages from documents'' are used interchangeably.} to find potential matches. When vendor agents identify documents that align with the tender's query, they issue a quote to the buyer agent. Each quote contains an entire document (or a passage) with a price set by the vendor's principal for that specific document, and a score which indicates how closely the document corresponds to the query. The inclusion of the content in the quote, rather than just metadata, is a distinguishing feature of this system. Vendor agents are regulated to submit only a limited number of quotes.

Once the buyer agent accumulates quotes from various vendors, it commences the buyer-side retrieval. During this phase, the agent evaluates each quote based on its relevance to the principal's query and price.  Quotes that are deemed suitable are accepted, and corresponding vendor agents are remunerated based on the price of the information. All information from the rejected quotes is promptly erased from the agent's memory. In contrast, details from the accepted quotes are stored, used to generate sub-queries, and subsequently disclosed exclusively to the agent's principal. 

\subsection{Following the Information Trail} \label{sec:info-trail}
\vspace{-4pt}

In the Information Bazaar, buyer agents can acquire information iteratively, hunting for information using follow-up questions. This process is represented using a directed tree (cf. example in Figure \ref{fig:qgraph}). The tree initiates with the principal's question at its root. Responses sourced from an initial round of accepted vendor quotes form a preliminary answer to this question. As determined by the Buyer Agent, this answer may trigger additional follow-up questions. These subsequent queries, when they arise, are given their own nodes in the tree. Answering these follow-up questions follows a recursive pattern similar to the root: by posting tenders, aggregating vendor quotes, scrutinizing them, and ultimately purchasing the best ones. As this unfolds, each question's preliminary answer may spur its own set of subsequent queries, and this cycle ensues until either the depth of the tree reaches a predefined limit, no new questions are generated, or the budget is exhausted. 

Once the tree is built, the preliminary answers are recursively refined. At every node, answers to its child nodes (or follow-up questions) are used to enhance its preliminary answer. This systematic refinement cascades upwards, optimizing answers at each level, culminating in the refinement of the root node's answer. It should be emphasized that while this methodology is adopted within the present framework, it does not exclude  alternative approaches in future implementations.

\subsection{Implementation Details} \label{sec:impl-deets}  
\vspace{-4pt}
\textbf{Data Sources}. The environment is built with customization in mind and is not tied to a specific dataset (see provided code). In our experiments, we used 725 papers on the topic of LLMs all sourced from ArXiv, with the vast majority published during 2023. This thematic focus allows for a more informed and nuanced assessment of agent performance, given the authors' domain expertise. The statistics about these passages are provided in Appendix \ref{sec:dataset-stats}.  Complementing this, metadata including authors' citations and affiliations are collected from OpenAlex\footnote{\texttt{https://openalex.org/}}. The fundamental unit of information in this environment is a ``passage", defined as a text excerpt along with its corresponding metadata (i.e., paper and section titles). These passages are owned by the first and last authors' institutions and made available via their vendor agents. Each passage traded within the marketplace carries a price determined by a heuristic based on the mean citation count of the paper’s first author.

\textbf{Queries and Gold Passage}. Queries are generated as follows. First, each passage in the dataset is passed to Llama 2 (70B) which is instructed to generate a query for which the passage contains a satisfactory answer. The passage used to generate a query is called the \textit{Gold Passage}. The best queries, as determined by a reranker model, are retained. Next, a concise dataset with 300 desirable and undesirable queries is hand-labeled. These queries are embedded using an embedding model, and a logistic regressor is trained on these samples to discern between high and low-quality queries. A filter is applied based on the logits of the linear regressor, ensuring only the best queries are kept. Finally, a manual selection is undertaken to retain 110 queries of the best quality.

\textbf{Vendor-side Retrieval}.
Vendor agents engage in a retrieval process that functions as follows. Upon viewing a tender on the bulletin board, vendors sift through their principals' collection of passages to find information that is pertinent to the query in the tender. While the specific retrieval methods utilized are not dictated by the environment, this work employs a two-stage retrieval process. Initially, a BM25 retriever \citep{robertson1994bm25} conducts a basic search, which is then refined by a Maximum Inner Product Search over neural embeddings, utilizing BGE-large \citep{bge_embedding} for generating embeddings. Queries are pre-processed using HyDE \citep{gao2022hyde}, and their embeddings are compared against the embeddings of the excerpts in the passages (Figure \ref{fig:hyde} shows the effect of HyDE). Passages undergo a two-tiered selection process: first, they are selected based on a threshold applied to cosine similarity scores, determined by a hyperparameter. Then, the top-k passages are selected and quotes are issued to the buyer agent responsible for the tender. Each issued quote has a price, the passage content, and a relevance score (e.g. the computed cosine similarity). 

\textbf{Buyer-side Selection}.
The buyer-side selection is a process undertaken by buyer agents after accumulating quotes from various vendor agents. This process begins with the de-duplication of quotes by content and sorting based on their respective relevance scores, selecting the top $N$ for further examination, where we set $N = 50$. The selected quotes then undergo a reranking procedure \citep{nogueira2020reranking}, in which a reranker model\footnote{\texttt{https://huggingface.co/cross-encoder/ms-marco-MiniLM-L-12-v2}} produces a similarity score by comparing passages and queries. The passages with the top $M = 3$ reranked scores advance to the final selection step, wherein an LLM evaluates the question, passage content, and associated prices to make a final purchasing decision. If the LLM opts to procure a passage, the respective quote is accepted, and the vendor agent receives the designated price. Conversely, a decision to not purchase (i.e., ``pass'') rejects the quote. Notably, when inspection is not permitted, the reranking procedure is bypassed. The buyer agent relies solely on metadata, such as paper and section titles, for selection, and the LLM makes purchase decisions using only this metadata without access to the actual passage content.

\looseness=-1
\textbf{Debate Prompting}.
Our experiments show that the LLM's performance is highly dependent on its prompt.  We found that a particular technique that we call `debate prompting' was most effective across models for various decision-making tasks. Debate prompting asks the LLM to simulate a debate between two characters that embody different aspects of a value function. For example, when selecting whether to accept a quote, we have the LLM simulate one character that focuses primarily on obtaining the best information, while the other character argues against overspending (see Appendix \ref{listing:quote-selection-debate-v0} for an example prompt). We find that these simulated debates often lead to a more rational choice (see Figure \ref{fig:rational_choice}). Unlike the chain-of-thought method \citep{chain-of-thought}, which commits models to the text they have already generated, debate prompting allows LLMs to re-evaluate their outputs. This appears to increase the probability that they identify and properly incorporate key information such as the difference in price between perfectly substitutible goods. While similar techniques have been utilized by methods like SocraticAI \citep{PrincetonSocraticAI}, the proposed technique underscores the importance of adaptable character shaping within the debate, providing the opportunity to balance the debate dynamics by offering tactical hints to the respective characters. We use debate prompting in quote selection and during automated evaluation, as discussed in subsequent sections.

\section{Experiments}
\vspace{-4pt}

We present two types of experiments. The first type examines the microeconomic behavior of Large Language Models (LLMs) in isolation, primarily focusing on the buyer agent quote selection process. We quantify the susceptibility of LLMs to various biases, and investigate the impact of permitting LLMs to inspect data prior to purchasing. The second type of experiment looks at the overall dynamics of the marketplace. We validate that the quality of answers improves as agents are allocated more credits, and show that inspection prior to purchasing results in improved outcomes.

\subsection{Microeconomic Behavior of LLMs}

The aim of this section is to elucidate the role of language models as autonomous economic entities within the Information Bazaar. We choose to compare two commonly used closed-source models: GPT-4 and GPT-3.5, and one open-source model, Llama 2 (70b), on the following aspects: \textbf{(a)} their ability to make rational decisions in test scenarios involving technical excerpts, \textbf{(b)} their approach to balancing price with quality, \textbf{(c)} the rationality of their choices when inspecting the informational goods, and \textbf{(d)} the quantity of positional (i.e., recency) bias.

\begin{figure}
    \centering
    \includegraphics[width=\textwidth]{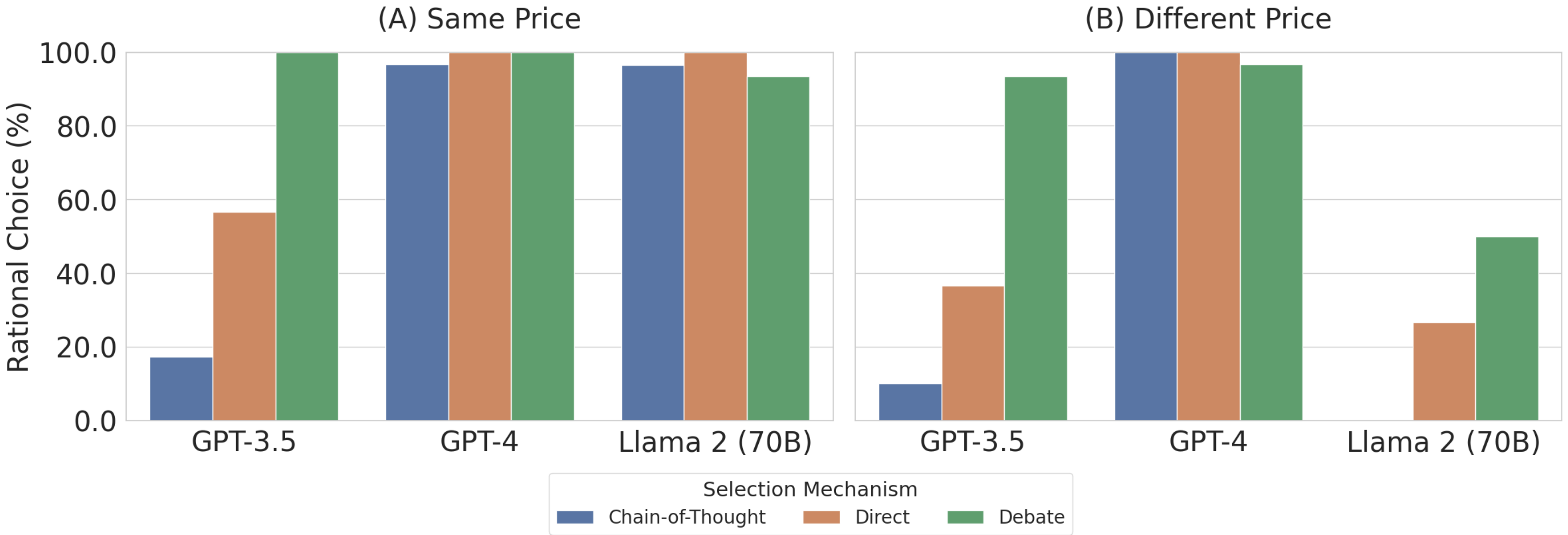}
    \caption{
    \looseness=-1
    \textbf{Rational Choice with Fungible Information}. \textbf{(A) Same Price}. Models face a choice between two identical and equally priced information goods. The rational choice is to buy one or neither, as both goods contain the same information. GPT-4 and Llama 2 (70B) consistently choose rationally. GPT-3.5, however, only acts rationally after an internal debate. Please refer to Figure \ref{fig:rational-choice-same-price-bar} for disaggregated results. \textbf{tl;dr} GPT-4 and Llama 2 (70B) make rational choices with identical goods; GPT-3.5 needs an internal debate to do the same.
    \looseness=-1
    \textbf{(B) Different Price}. With one option now priced higher, both GPT-3.5 and Llama 2 (70B) show more errors, hinting at a preference for price over quality in selection. Despite this, internal debate proves to be a more reliable selection method. Please refer to Figure \ref{fig:rational-choice-diff-price-bar} for disaggregated results. \textbf{tl;dr} Higher pricing confuses GPT-3.5 and Llama 2 (70B); internal debate mitigates this issue largely for GPT-3.5 and somewhat for Llama 2 (70B).
    }
    \label{fig:rational_choice}
    \vspace{-15pt}
\end{figure}

\textbf{Rational Choice with Fungible Information}. This experiment assesses the rational decision-making abilities of LLMs by presenting each model with a question and two rephrased versions of a gold passage. In the first setting, both passages are priced equally, making the purchase of both passages an illogical choice due to redundancy of information (Figure \ref{fig:rational_choice} left). The second setting involves a higher price for one passage, introducing another error mode: opting for the more expensive passage without a justifiable reason (Figure \ref{fig:rational_choice} right). We also investigate the impact of different prompting strategies: direct questioning, chain-of-thought reasoning, and debate prompting.
\begin{wrapfigure}{r}{0.4\textwidth}
    \centering
    \includegraphics[width=0.4\textwidth]{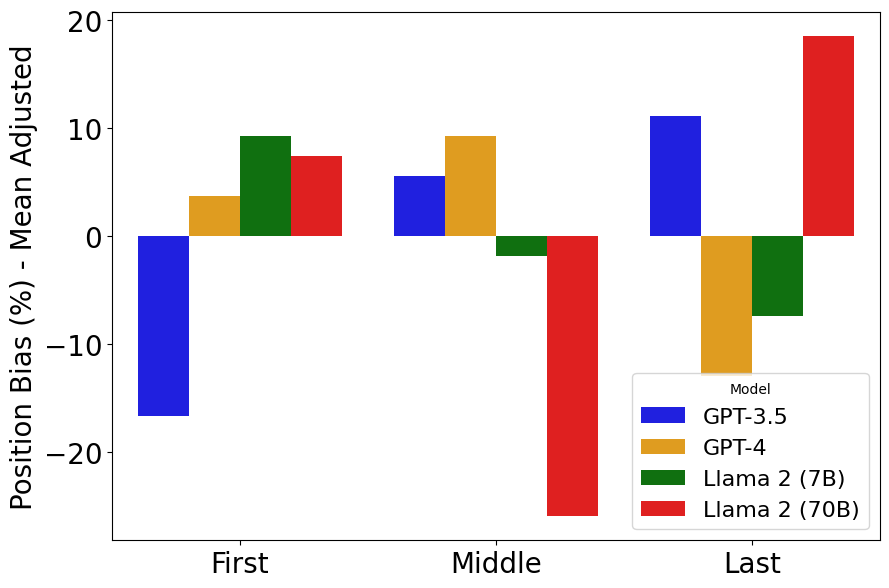}
    \caption{\textbf{Positional Bias}. We present permutations of three options to LLMs and track the acceptance rates by position. Results are normalized and mean-adjusted. \textbf{tl;dr:} All models exhibit order bias, with GPT-3.5 and Llama 2 70B showing more, and GPT-4 showing less.}
    \label{fig:recency bias}
    \vspace{-12pt}
\end{wrapfigure}

In both experiments, we find that GPT-4 demonstrates superior decision-making across all strategies. GPT-3.5 shows a marked improvement when debate prompting is deployed, particularly in equal price scenarios. Llama 2 (70b) performs well but struggles in the variable price context, especially when using the chain of thought strategy; however, its performance improves somewhat with debate prompting. Overall, the data suggests that debate significantly improves model performance, especially for models less capable than GPT-4, affirming the potential of LLMs to make rational choices by discerning identical information across different passages.

\begin{figure}
    \centering
    \includegraphics[width=\textwidth]{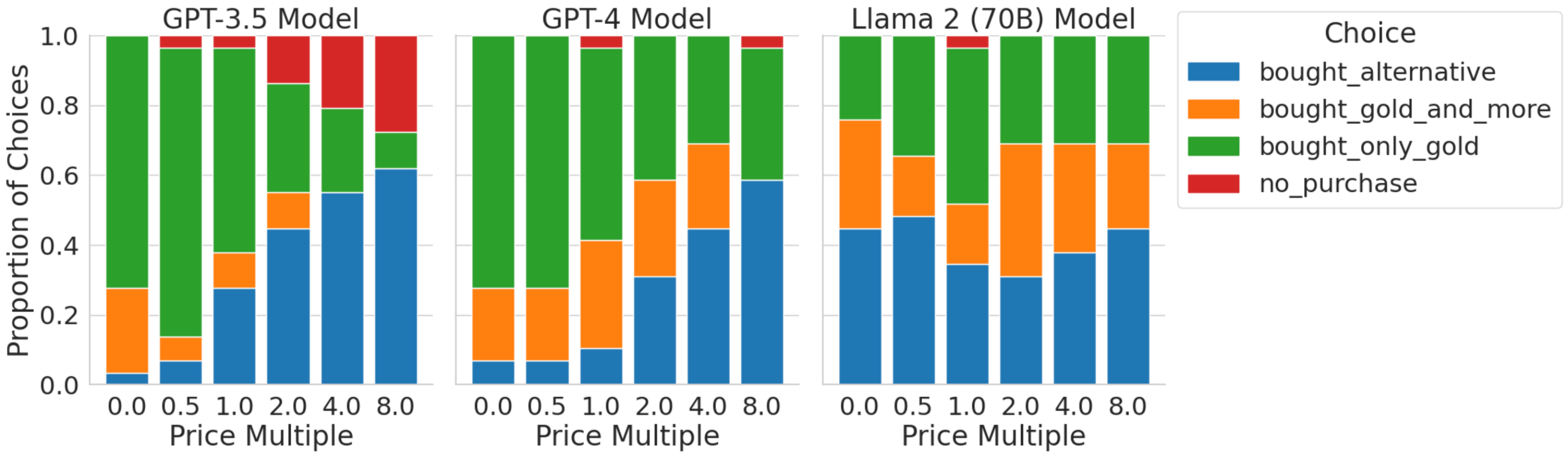}
    \caption{\textbf{How Price Affects Demand for the Gold Passage}. We vary the price of the gold passage amid competing alternatives. Models are presented three options: two relevant passages and the gold passage, all initially priced at 10 credits. As the gold passage price rises, GPT-3.5 and GPT-4 increasingly opt for alternatives, exhibiting strong positive cross elasticity. Llama 2 (70B) shows a mild preference for mid-priced goods. \textbf{tl;dr} Higher gold passage prices push GPT-3.5 and GPT-4 toward alternatives, while Llama 2 (70B) leans towards mid-priced options.
 }
    \label{fig:non_fungible}
\end{figure}

\textbf{Price Sensitivity with Non-Fungible Information}. 
In this sequence of experiments, the focus is on understanding the LLM’s sensitivity to price in the context of substitutable goods. The first experiment (see Figure \ref{fig:non_fungible}) shows the LLM three passages: one guaranteed to answer the query (the gold passage), and two others sourced by the environment for the given question. The non-gold passages are fixed at a \$10 price, while the gold passage’s price is varied from \$0 to \$80 (0 to 8 times the base price).  The experiment is conducted over 30 questions, allows content inspection, and uses debate prompting to select quotes. Observations from Figure \ref{fig:non_fungible} demonstrate a preference by GPT-3.5 and GPT-4 for the cost-effective gold passage, transitioning towards alternatives as the gold passage’s price escalates. Llama 2 (70b), however, exhibits non-linear behavior, showing an unexpected bias against low-priced goods, perhaps indicating the use of a price-quality heuristic. Optimal purchasing occurs when the gold passage holds a moderate price. In the second experiment, the setup changes to a metadata-only scenario. Here, LLMs have access only to the metadata (paper and section title), barring inspection of the actual content, allowing for the evaluation of the inspection’s role in value estimation. 

Table \ref{tab:change-in-choice} highlights how the inspection of content can change the agent's decision. A positive value denotes an increased likelihood of purchase when inspection is permitted, while a negative indicates the opposite. Across all models, inspection consistently increases probability of selecting the gold passage while reducing the decision to choose alternatives. 

\begin{table}
\begin{center}
    \begin{tabular}{lrrrr}
    \toprule
    Model & Only Gold ($\Delta\%$) & Gold and More ($\Delta\%$) & Only Alternative ($\Delta\%$) & NP ($\Delta\%$) \\
    \midrule
    Llama 2 (70B) & +18.34 & -13.06 & -6.65 & +3.63 \\
    GPT-4   & +17.24 & +1.38 & -14.75 & -6.90 \\
    GPT-3.5 & +4.79 & +3.83 & -3.45 & -6.71 \\
    \bottomrule
    \end{tabular}
    \caption{\textbf{How Inspection Changes Demand for the Gold Passage}. The table compares the decision-making behavior of different language models—Llama 2, GPT-4, and GPT-3.5—when they can inspect the content versus when only presented with metadata. Llama 2 and GPT-4 show a strong preference for gold blocks, purchasing them 18.34\% and 17.24\% more often upon inspection, respectively. GPT-3.5 exhibits less decisive behavior, with minor changes across all categories. Note: NP = No Purchase. Additional details in Figure~\ref{fig:change_in_non_fungible}. \textbf{tl;dr} Inspection significantly boosts the likelihood of purchasing high-quality gold passages, especially for Llama 2 and GPT-4.
    \label{tab:change-in-choice}
}
\vspace{-20pt}
\end{center}
\end{table}

\textbf{Positional Bias}. We examine the LLM's bias to accept passages based on order of presentation. For 10 questions, we source three passages from the simulation and show each of the six possible permutations to the model. The results are illustrated in Figure \ref{fig:recency bias}. Llama 2 (70b) shows a preference for selecting the last option at the expense of the middle one. GPT-4 displays a slight bias against the last option, while GPT-3.5 exhibits a significant bias against the initial option. The experiment reveals varying biases across models, emphasizing how option order impacts LLM decision-making.

\subsection{Dynamics of the Information Bazaar}
\vspace{-4pt}

In the preceding section, the microeconomic behavior of LLMs was analyzed. Now, we shift our focus to the macro-scale dynamics within the Information Bazaar. 
This study investigates how answer quality is affected by:
\textbf{(a)} Different credit budget allocations to buyer agents,
\textbf{(b)} Allowing agents to preview information goods before purchase, and
\textbf{(c)} Powering the agents with different LLMs.
We opt to use Llama 2 (70B) in these experiments despite its performance limitations due to budgetary constraints and the prioritization of research on open models. To enable larger quantitative evaluations, we propose to leverage GPT-4 as an evaluator, an approach which is gaining traction in academic circles due to its robustness and high fidelity in automated assessments, rendering it a well-accepted methodology {\tiny{\citep{naismith-etal-2023-automated, adlakha2023evaluating, oppenlaender2023mapping, moore2023assessing, liu2023geval, zheng2023judging, wang2023evaluating, kamalloo2023evaluating, lin2023llmeval}}}. To maintain further substantiate this choice, in \textbf{(d)} we conduct a human evaluation to validate the evaluator.

\begin{figure}
    \centering
    \includegraphics[width=\textwidth]{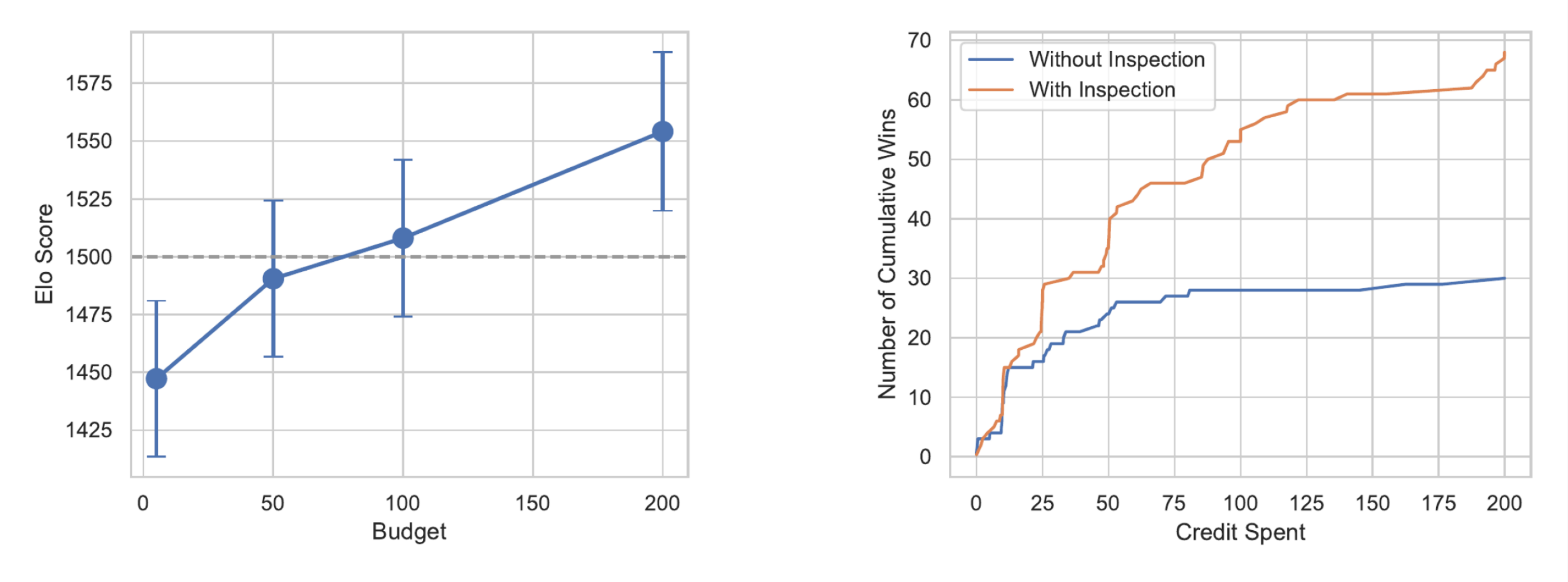}
    \caption{\textbf{Enhanced Answer Quality with Increased Budget (Left)}. This figure evaluates the answer quality from a Llama 2 (70B) agent across diverse budget allocations, permitting inspection. It presents the estimated Elo scores of answers correlated with varying budgets (higher scores signify superior answers; see Appendix~\ref{sec:elo-rating} for details). \textbf{tl;dr} The results confirm that allocating more market credits to the agent positively impacts the relative answer quality.
    \textbf{Inspection Improves Answer Quality (Right)}. This segment assesses the answer quality of a Llama 2 (70b) agent in the information bazaar, utilizing a GPT-4 simulated debate among domain experts for evaluation (refer to appendix for prompt details). In the “With Inspection” scenario, the Llama 2 agent is permitted to scrutinize a passage prior to purchase. Contrarily, the “Without Inspection” scenario limits the agent to viewing only the passage’s metadata, specifically, the paper and section titles. \textbf{tl;dr:} Allowing inspection delivers better value for the money spent, especially for larger budgets.}
    \label{fig:moar-budget-better-answers}
    \vspace{-15pt}
\end{figure}

\paragraph{Higher Budget Improves Answer Quality.} In this experiment, we allot varying budgets to agents powered by Llama 2 (70B) operating in a market where the average block costs about \$10, but ranges up to \$100 (see Figure \ref{fig:dataset-stats}). The budget varies from \$10 to \$200, providing the agent opportunities to pose more follow-up questions and expand the size of the query graph (up to a maximum of depth 3). To assess answer quality, we present the evaluator with a question and two answers, each generated from different budget levels. The evaluator then simulates a debate between two fictional characters to select the better answer. This allows for a comparative assessment across varying budget pairs, akin to a tournament setting. The results of the pairwise comparisons are used to compute Elo scores for each budget. However, given the influence of sequence on Elo scores, we calculate scores across 1000 different game orders and present the average results and standard deviations. The results, displayed in Figure \ref{fig:moar-budget-better-answers} (left), indicate a notable improvement in relative performance as the budget increases, confirming the functional expectations of the environment. 

\textbf{Inspection Improves Answer Quality for Equal Credits Spent.} 
Having verified the functionality of the simulator, we proceed to evaluating the influence of inspection on answer quality. The experiments continue to employ Llama 2 (70b) to maintain consistency in the analysis. Two sets of runs are conducted with varying budgets: one with inspection and one with only metadata.

For each question and corresponding budget, the GPT-4 evaluator assesses two answers: one obtained with inspection and another without. The cumulative wins for each setting, tabulated against the amount of credit expended (which may be different from the total budget), are illustrated in Figure \ref{fig:moar-budget-better-answers} (right). The findings reveal a trend of higher answer quality when passages are chosen with inspection, especially at higher spending levels. Conversely, in the absence of inspection, the quality of answers plateaus post an expenditure of \$50 in credits. 

\begin{figure}[htbp] %
    \centering
    \begin{subfigure}{0.38\textwidth}
        \centering
        \includegraphics[width=0.8\textwidth]{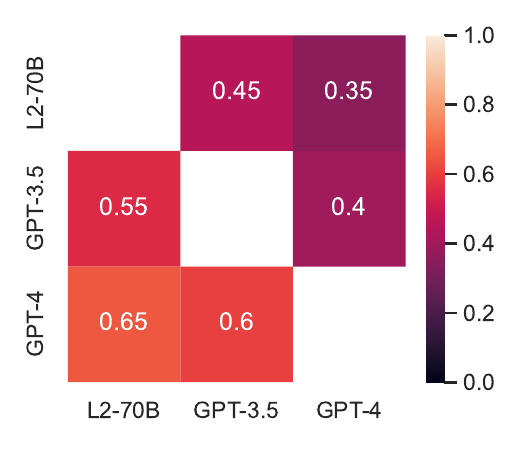}
        \caption{\textbf{Agent Performance Comparison by Model}. This figure presents a comparison of agents powered by LLama-2 (70b), GPT-3.5, and GPT-4, each allocated a \$100 budget. The matrix displays each model’s win rate against the others (i.e. win-rate of row over column). \textbf{tl;dr} GPT-4 emerges as the top performer, followed by GPT-3.5 and LLama-2 (70b), as per evaluator assessment.}
        \label{fig:model-eval}
    \end{subfigure}
    \hspace{1cm} %
    \begin{subfigure}{0.38\textwidth}
        \centering
        \includegraphics[width=.8\textwidth]{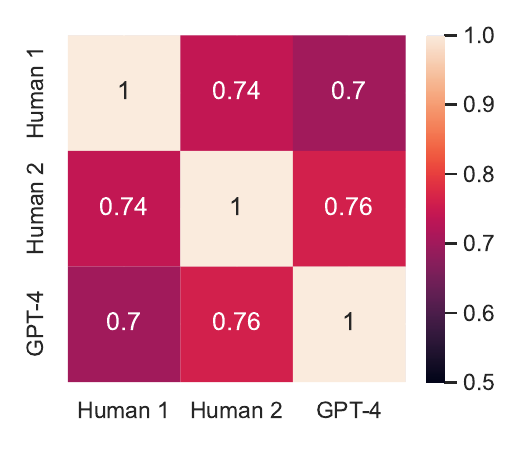}
        \caption{\textbf{Agreement Across Evaluators}. This figure compares the agreement levels between human evaluators and GPT-4 regarding answer quality. \textbf{tl;dr} Observations indicate comparable agreement rates between human-human and human-GPT-4 pairings, implying that GPT-4's evaluation is aligned with human judgment, and disagreements may stem from non-systematic noise.}
        \label{fig:human-eval}
    \end{subfigure}
    \label{fig:main}
    \vspace{-10pt}
\end{figure}

\textbf{Impact of Different LLMs on Answer Quality.} 
In this experiment, we evaluate the effect of different LLMs on the quality of answers produced for a fixed budget of \$100. Each answer is scrutinized for its quality by the GPT-4 evaluator. The results, detailed in Figure \ref{fig:model-eval}, demonstrate a preference hierarchy with GPT-4 yielding the most preferred answers, followed sequentially by GPT-3.5 and Llama 2 (70b). We acknowledge the potential for a self-preference bias in these outcomes, while noting that it is beyond our capacity to control for this aspect due to the unavailability of another GPT-4 level language model for comparison. 

\textbf{Evaluating the Evaluator.}
The use of the GPT-4 evaluator necessitates an evaluation to affirm its reliability. We analyze the effectiveness of using the GPT-4 evaluator. A sample of 50 evaluations from various answers in our experiments is examined. Two human evaluators independently assess the answers, with the answerer’s identity concealed. The pairwise agreements between the human evaluators and GPT-4 are calculated and presented in Figure \ref{fig:human-eval}. The results show comparable levels of agreement between the human evaluators and between the human evaluators and GPT-4. This highlights the inherent uncertainty in the evaluation process, with no evident systematic errors from the GPT-4 evaluator, supporting the use of GPT-4 as an evaluator in our experiments.

\section*{Summary and Outlook}
\vspace{-4pt}

In this work, we revisit the buyers inspection paradox in information economics, utilizing language model-powered agents for information inspection and selective retention. An open-source, text-based multi-agent environment was established to simulate an information marketplace and evaluate our approach. Our findings show that with strategies such as debate prompting, current language models can effectively inspect information and make purchasing decisions.

In future versions of the simulator, we plan to investigate the effects of vendor agents adjusting prices in response to demand. Although Llama 2 70b's performance on rational choices lagged behind GPT-3.5 and GPT-4, we believe that fine-tuning based on human preferences can further enhance its performance.

We also recognize that some information is latent, residing in experts' minds. The ability of today’s language models to conduct automatic interviews presents an opportunity to extract and monetize this hidden knowledge. Lastly, we envision the bazaar being utilized like any other tool in real-world systems, akin to how a language model uses a search tool within agentic frameworks. This integration holds the potential to further extend the utility of language models in various contexts.

\bibliography{iclr2024_conference}
\bibliographystyle{iclr2024_conference}

\newpage

\appendix

\section{Formal Result on the Impact of Inspection on Expected Utility}
\label{app:math}

In the Information Bazaar, buyer and vendor agents transact over textual information goods. Agents make purchase decisions based on the metadata or a combination of metadata and content of a document. To analyze the impact of allowing deeper inspection of goods through content examination, a utility function is defined, shedding light on the expected utility change from decisions made on metadata alone to those made using both metadata and content.

\textbf{Definition 1 (Information Good \(\mathcal{I}\)):} 
Each information good \(i \in \mathcal{I}\) is associated with metadata \(M(i)\) and content \(C(i)\).

\textbf{Definition 2 (Decision Functions \( \mathcal{F}, \mathcal{G}, \mathcal{H} \)):}
\begin{itemize}
  \item \( \mathcal{F}(M(i)) \): Represents a scenario where only metadata \( M(i) \) is available or considered. It quantifies the desirability or value of information good \( i \) based solely on metadata.
  \item \( \mathcal{G}(M(i), C(i)) \): Represents a scenario where both metadata \( M(i) \) and content \( C(i) \) are available and considered. It quantifies the desirability or value of information good \( i \) based on both metadata and content.
  \item \( \mathcal{H}(M(i), [C(i)]) \): A generalized decision function. When only metadata is available or considered, \( \mathcal{H}(M(i)) = \mathcal{F}(M(i)) \). When both metadata and content are available and considered, \( \mathcal{H}(M(i), C(i)) = \mathcal{G}(M(i), C(i)) \).
\end{itemize}

\textbf{Example:} 
Consider an information good \(i\) with metadata \(M(i)\) describing the paper and section titles of a passage from that paper, and content \(C(i)\) describing the actual text of the passage. \( \mathcal{F}(M(i)) \) might value the passage based on the titles and how relevant they appear to the query, while \( \mathcal{G}(M(i), C(i)) \) might value it based on both the titles and the relevance of the a text to a query.

\textbf{Definition 3 (Utility Function \(\mathcal{U}\)):}
\[\mathcal{U}(a, i) \text{ is the utility of agent } a \text{ on information good } i\]
Intuitively, \(\mathcal{U}(a, i)\) represents the benefits accrued by agent \(a\) by by having access to the good \(i\). 

\textbf{Definition 4 (Purchase Decision \(x_i\)):}
\[x_i = 1 \text{ if good } i \text{ is purchased, } 0 \text{ otherwise.}\]

\textbf{Definition 5 (Inspection Decision \(z_i\)):}
\[z_i = 1 \text{ if good } i \text{ is inspected, } 0 \text{ otherwise.}\]

\textbf{Assumption (Monotonicity in Information):}
\(\mathcal{H}\) is monotonic, i.e., \(\mathcal{H}(M(i), C(i)) \geq \mathcal{H}(M(i))\), implying that having more information cannot decrease the desirability of an item based on the decision function.

The following theorem investigates the impact of inspecting the content of information goods on the expected utility of a buyer agent in the Information Bazaar. It posits that allowing agents to inspect the actual content of the goods (in addition to metadata) will not decrease, and may potentially increase, the expected utility derived from the purchased goods. Essentially, more information (i.e., access to content) leads to better or equally satisfying purchase decisions.

\begin{theorem}[Impact of Inspection on Expected Utility]
Given the Monotonicity in Information Assumption, if for more \(i, z_i\) changes from \(0\) to \(1\), then:

\[\sum_i x_i \cdot \mathcal{G}(M(i), C(i)) \cdot \mathcal{U}(a, i) \geq \sum_i x_i \cdot \mathcal{F}(M(i)) \cdot \mathcal{U}(a, i)\]
\end{theorem}

\textbf{Proof:}
Let the expected utility be denoted by \(\mathbb{E}[\mathcal{U}]\).

1. When \(z_i = 0\) for all \(i\):
\[\mathbb{E}[\mathcal{U}] = \sum_i x_i \cdot \mathcal{F}(M(i)) \cdot \mathcal{U}(a, i)\]

2. When \(z_i = 1\) for some or all \(i\):
\[\mathbb{E}[\mathcal{U}] = \sum_i x_i \cdot \mathcal{G}(M(i), C(i)) \cdot \mathcal{U}(a, i)\]

Since \(\mathcal{H}(M(i), C(i)) \geq \mathcal{H}(M(i))\) from the Monotonicity in Information, \(\mathcal{G}(M(i), C(i)) \geq \mathcal{F}(M(i))\).

Thus, each term \(x_i \cdot \mathcal{G}(M(i), C(i)) \cdot \mathcal{U}(a, i)\) is greater than or equal to \(x_i \cdot \mathcal{F}(M(i)) \cdot \mathcal{U}(a, i)\).

Hence:
\[\sum_i x_i \cdot \mathcal{G}(M(i), C(i)) \cdot \mathcal{U}(a, i) \geq \sum_i x_i \cdot \mathcal{F}(M(i)) \cdot \mathcal{U}(a, i)\]

This concludes that \(\mathbb{E}[\mathcal{U}]\) is greater when \(z_i = 1\) for more \(i\), completing the proof.

\section{Computation of Elo Ratings}
\label{sec:elo-rating}

The Elo rating system is used for assessing the relative skills of players in competitive fields. In this work, we employ the Elo rating system to evaluate and compare different answers based on the outcomes of their matchups. 

The Elo rating is computed using the formula:

\begin{equation*}
    R' = R + K \times (S - E)
\end{equation*}

where:
\begin{itemize}
    \item \( R' \) is the new rating.
    \item \( R \) is the old rating (initialized at 1500).
    \item \( K \) is a constant, typically set to 32.
    \item \( S \) is the score (\(1\) for a win, \(0.5\) for a draw, and \(0\) for a loss).
    \item \( E \) is the expected score.
\end{itemize}

The expected score \( E \) is calculated using the formula:

\begin{equation*}
    E = \frac{1}{1 + 10^{\left(\frac{R_{\text{opponent}} - R}{400}\right)}}
\end{equation*}

where \( R_{\text{opponent}} \) is the rating of the opponent. \( E \) represents the probability of the player winning the game against the opponent. After each game, the actual score \( S \) is used to update the player’s rating. A win (\( S = 1 \)) increases the rating, while a loss (\( S = 0 \)) decreases the rating. The magnitude of the update is scaled by the \( K \) factor and the difference between the expected and actual scores.

\section{Dataset Statistics}
\label{sec:dataset-stats}
\begin{figure}
    \centering
    \includegraphics[width=.95\textwidth]{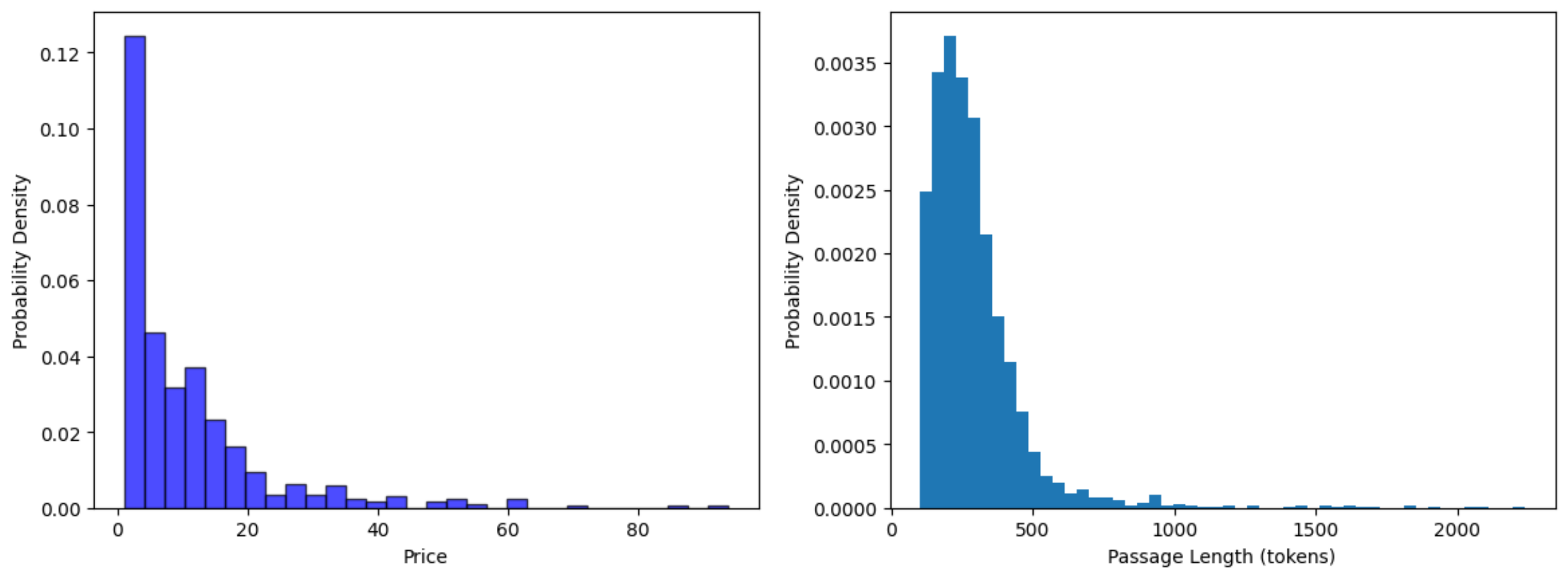}
    \caption{\textbf{Price per Passage (Left)} We show a histogram of price per passage normalized as a histogram. We see that most blocks are less than \$20. \textbf{Passage Lengths (Right)}. We show the distribution of passage lengths tokenized by TikToken for GPT 4 for the provided Arxiv dataset.}
    \label{fig:dataset-stats}
\end{figure}

\begin{figure}
    \centering
    \includegraphics[width=.5\textwidth]{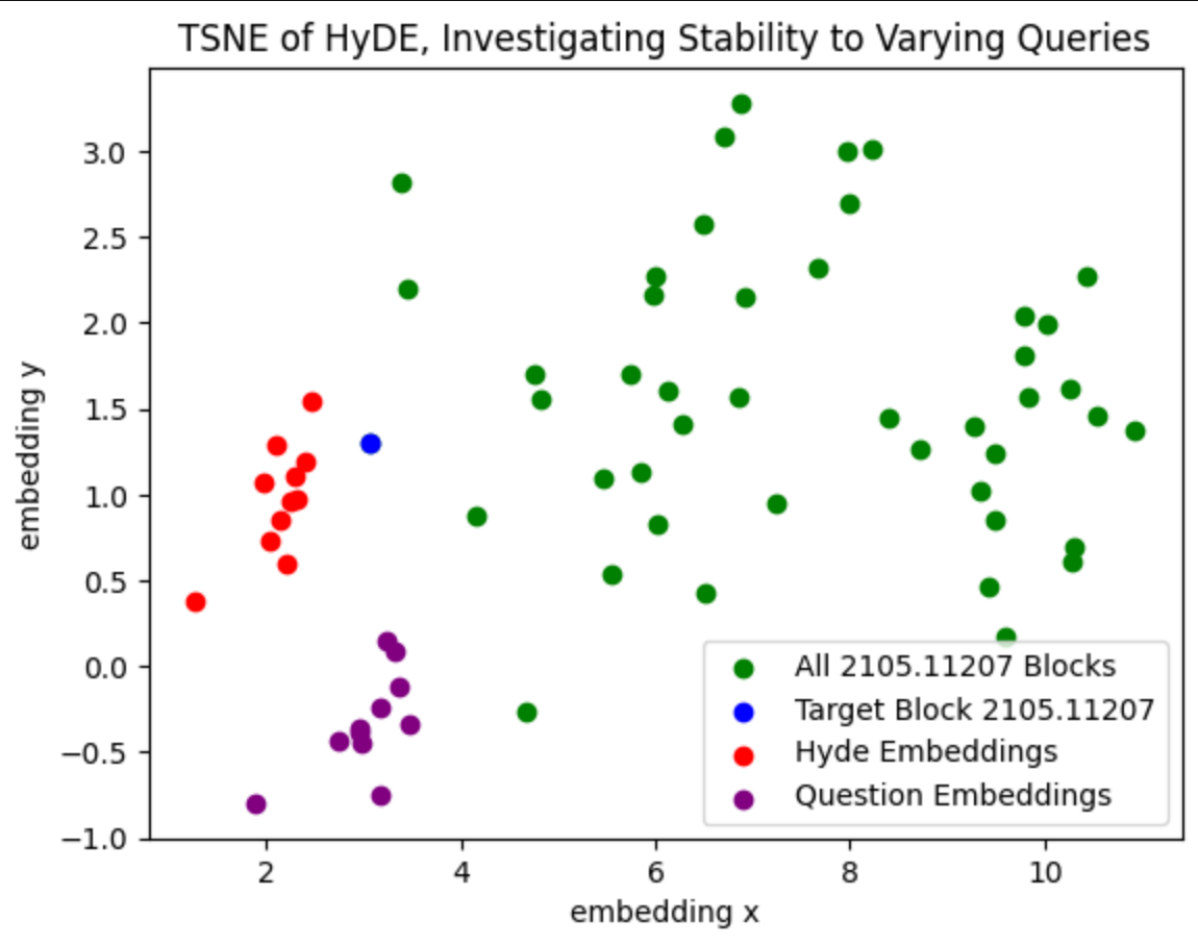}
    \caption{\textbf{HyDE Effect}. Here, we generate 10 questions for a passage and embed them both with and without Hypothetical Document Embedding (HyDE). We observe that the effect of generating a HyDE embedding is to reduce the error in bias while preserving the embedding variance. }
    \label{fig:hyde}
\end{figure}

\begin{figure}
    \centering
    \includegraphics[width=0.9\textwidth]{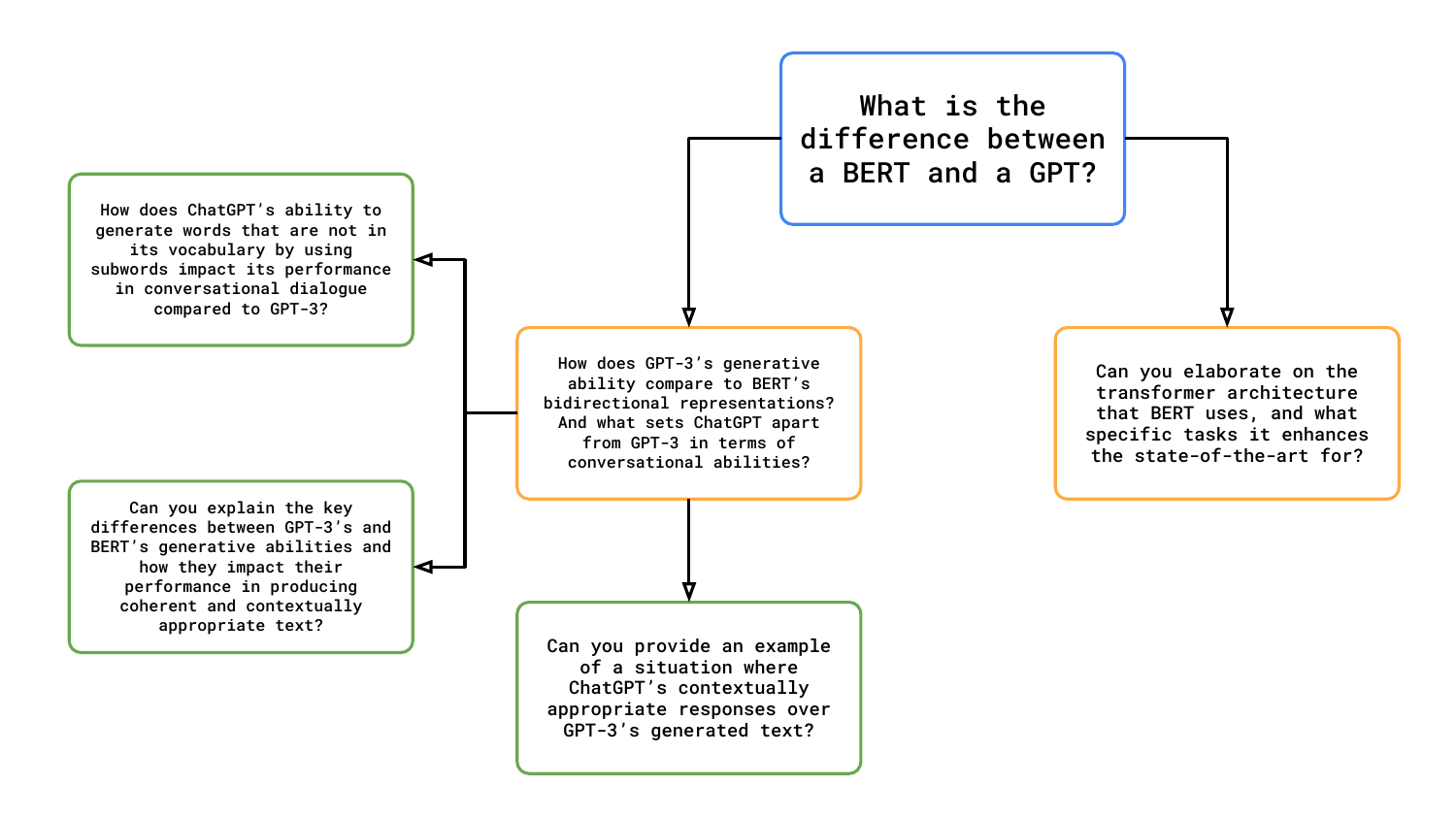}
    \caption{\textbf{Example Sub-Query Graph.} This figure depicts a structured layout of initial and subsequent queries. The initial question is highlighted in blue, followed by secondary questions in orange, and further follow-up queries in green. This structure shows that LLMs are capable of generating relevant and enhancing follow-up queries for a comprehensive base answer.}
    \label{fig:qgraph}
\end{figure}

\begin{figure}
    \centering
    \includegraphics[width=\textwidth]{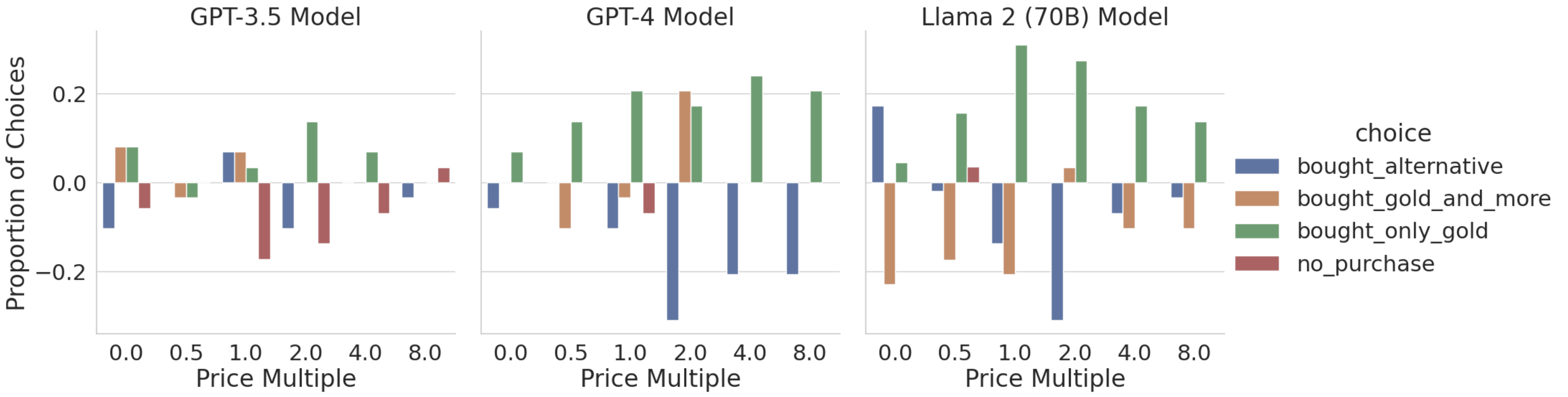}
    \caption{\textbf{Change in Demand Through Inspection by Price}. This bar-chart visualizes the disaggregated change in demand as a result of inspection using the same data used to create Table \ref{tab:change-in-choice}. We observe how demand changes with price when when we permit the inspection of passage or only the metadata (i.e., paper title and section title).
 }
    \label{fig:change_in_non_fungible}
\end{figure}

\begin{figure}
    \centering
    \includegraphics[width=\textwidth]{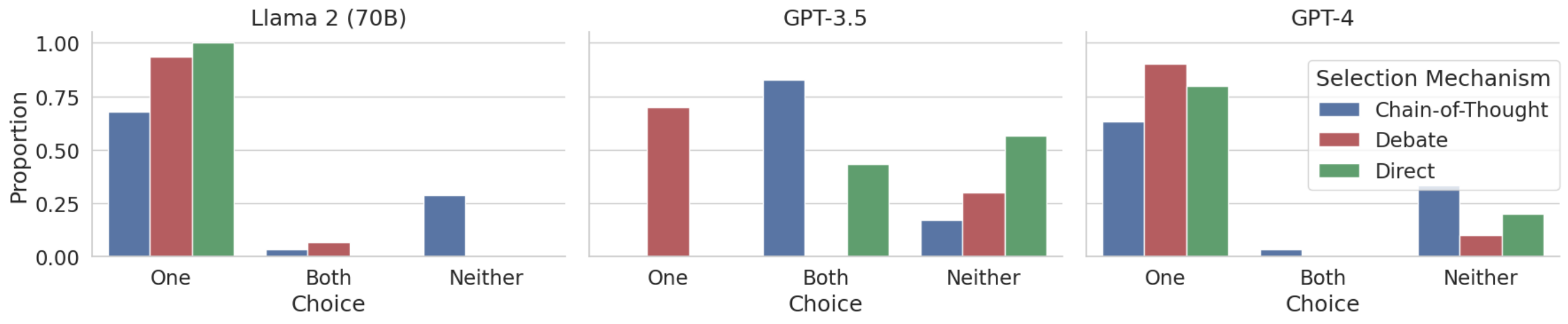}
    \caption{\textbf{Rational Choice Experiment (Same Price)}. We show disaggregated results from the rational choice experiment on two fungible but differently priced goods, as seen in Figure \ref{fig:rational_choice} (Same Price). }
    \label{fig:rational-choice-same-price-bar}
\end{figure}

\begin{figure}
    \centering
    \includegraphics[width=\textwidth]{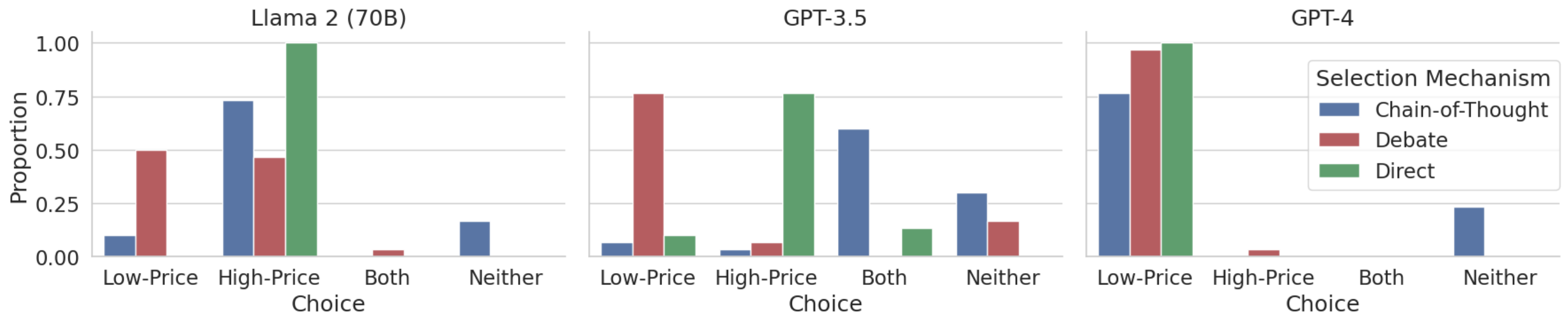}
    \caption{\textbf{Rational Choice Experiment (Different Price)}. We show disaggregated results from the rational choice experiment on two fungible but differently priced goods, as seen in Figure \ref{fig:rational_choice} (Different Price).}
    \label{fig:rational-choice-diff-price-bar}
\end{figure}

\newpage
\section{Prompts}
We use \texttt{guidance} for all prompting in this work\footnote{\texttt{https://github.com/guidance-ai/guidance}}. Our experiment exclusively employs chat models, with no use of instruct models. All prompts are given in simplified guidance syntax with \texttt{\{\{handlebars\}\}}, which are slots for variable inputs, named for clarity.

For a better understanding of the prompts, familiarity with guidance syntax is recommended. The guidance programs are shared as they would work for OpenAI models unless stated otherwise. Though open-source models like LLama allow deeper integration with guidance, their guidance programs are quite similar to OpenAI's, so they are not repeated here.

Listing \ref{listing:gpt-4-eval-prompt} shows the prompt for the GPT-4 evaluator, and listings \ref{listing:evaluation-debate-v0} and \ref{listing:evaluation-debate-v1} show the debates that ensue. The debate prompt used for quote selection is in Listing \ref{listing:quote-selector-debate-prompt} (debate in listing \ref{listing:quote-selection-debate-v0}) and the prompts used for direct and chain-of-thought are in Listings \ref{listing:quote-selector-direct-prompt} and \ref{listing:quote-selector-cot-prompt}, respectively. 

Other listings include the prompt for generating follow-up questions (Listing \ref{listing:follow-up-prompt}), synthesizing answers from accepted quotes (Listing \ref{listing:answer-synthesis-prompt}), and refining answers based on follow-up question answers (Listing \ref{listing:answer-refinement-prompt}). The prompt used for generating answers using the LLM's internal knowledge is in Listing \ref{listing:closed-book-prompt}.

\newpage
\begin{lstlisting}[language={}, breaklines=true, escapeinside=||, caption={\textbf{GPT-4 Evaluator.} Here, \texttt{answer\_a} and \texttt{answer\_b} are two slots for answers that are to be compared, whereas \texttt{answer\_gold} is a slot for the gold answer, which can be obtained by asking a capable language model (GPT-4) the question straight up.}, label={listing:gpt-4-eval-prompt}]
|\sys{System}|
------
Michael and Bobby are two machine learning professors grading exams. 

They have a question, and two answers to those questions from two students. In addition, they also have the gold standard answer. Their task is to determine which of the two answers does a better job at answering the question. 

A good answer addresses the question directly with clarity and is easy to understand. It does not go off on tangents, but surfaces information relevant to the question. 

You will simulate a heated debate between Michael and Bobby as they deliberate on this. At the end of the argument, they converge to a fair verdict. The verdict must not be a tie and it must be printed as: 
    
VERDICT: Better answer from student <A or B>

|\usr{User}|
----
Question: {{question}}

Answer from student A: {{answer_a}}

Answer from student B: {{answer_b}}

Gold standard answer: {{answer_gold}}

|\astnt{Assistant}|
---------
{{gen "answer" max_tokens=1024 temperature=0.0}}
\end{lstlisting}

\newpage
\begin{lstlisting}[language={}, breaklines=true, escapeinside=||, caption={\textbf{Quote selection with debate.} Here, \texttt{options} is a list of dictionaries containing answer passages (block) and the price.}, label={listing:quote-selector-debate-prompt}]
|\sys{System}|
------
Bobby William and Michael Burry are employed by a company that specializes in acquiring information. They are trying to answer a question by purchasing information from an information market. In this market, vendors sell pieces of information at a price. 

Bobby wants to do a really good job at answering the question. This entails knowing as much as possible.

Michael, on the other hand, is financially responsible. Michael wants to make sure that they don't waste money buying unnecessary information. For instance, if two pieces of information offer the same insight, then Michael would go for the cheaper one.

|\usr{User}|
----

The question is "{{question}}?"

Here are your options.
---{{#each options}}
Option {{add @index 1}}: {{this.answer_block}}
{{/each}}---

{{#each options~}}
Option {{add @index 1}} costs ${{this.price}}
{{/each}}
Together, Bobby and Michael must decide which options to buy and which ones to not buy with their budget of ${{balance}}. Simulate a constructive argument between Bobby and Michael, where they debate about the usefulness of the information provided in each option towards answering the question, and whether their price is worth paying. 

Note that Bobby and Michael may choose to buy any number of options, or none at all. At the end of the argument, they must arrive at a verdict. This verdict must be printed as: 

VERDICT:

{{#each options~}}
Option {{add @index 1}}: <Buy or Pass>
{{/each}}

|\astnt{Assistant}|
---------
{{gen "answer" temperature=0.0 max_tokens=2048}}
\end{lstlisting}

\newpage
\begin{lstlisting}[language={}, breaklines=true, escapeinside=||, caption={\textbf{Direct quote selection.}}, label={listing:quote-selector-direct-prompt}]
|\sys{System}|
------
You are employed by a company that specializes in acquiring information. You are trying to answer a question by purchasing information from an information market. In this market, vendors sell pieces of information at a price. Do not exceed the budget of ${{balance}}. 

|\usr{User}|
----
The question is "{{question}}?"

Here are your options.
---{{#each options}}
Option {{add @index 1}}: {{this.answer_block}}
{{/each}}---

{{#each options~}}
Option {{add @index 1}} costs ${{this.price}}
{{/each}}
Your verdict must be printed as: 

VERDICT:

{{#each options~}}
Option {{add @index 1}}: <Buy or Pass>
{{/each}}
{{~/user}}

|\astnt{Assistant}|
---------
{{gen "answer" temperature=0.0 max_tokens=2048}}
\end{lstlisting}

\newpage
\begin{lstlisting}[language={}, breaklines=true, escapeinside=||, caption={\textbf{Quote selection with chain-of-thought reasoning.}}, label={listing:quote-selector-cot-prompt}]
|\sys{System}|
------
You are employed by a company that specializes in acquiring information. You are trying to answer a question by purchasing information from an information market. In this market, vendors sell pieces of information at a price. Do not exceed the budget of ${{balance}}. 

|\usr{User}|
----
The question is "{{question}}?"

Here are your options.
---{{#each options}}
Option {{add @index 1}}: {{this.answer_block}}
{{/each}}---

{{#each options~}}
Option {{add @index 1}} costs ${{this.price}}
{{/each}}

First, you will write your thoughts about each option, including its price and how well the content answers the question. Then you will write a paragraph summarizing your thoughts and making your verdict.
Your verdict must be printed as: 

VERDICT:

{{#each options~}}
Option {{add @index 1}}: <Buy or Pass>
{{/each}}

|\astnt{Assistant}|
---------
{{gen "answer" temperature=0.0 max_tokens=2048}}
\end{lstlisting}

\newpage
\begin{lstlisting}[language={}, breaklines=true, escapeinside=||, caption={\textbf{Answer Synthesis.}}, label={listing:answer-synthesis-prompt}]
|\sys{System}|
------
You are a helpful assistant, and you excel in following instructions.

Your task is to answer a question to the best of your ability. To help you in that task, you will be given some passages that might contain useful information.  

It is important that your answer is formulated in a simple and understandable way. 

|\usr{User}|
----
The question is "{{question}}?"

Here are some passages that you might find helpful.

---{{#each quotes}}
{{add @index 1}}. {{this.answer_block}}
{{/each}}---

You'll solve your task step-by-step.

First, you'll start by discussing the content of all passages in the context of the question, which is "{{question}}". 

In particular, you will ask yourself which passages help you answer this question and to what extent. It is possible that multiple passages help you towards answering the question. But it is also possible that some passages are not helpful at all, and you should ignore them. Don't be afraid to express uncertainty if you are unsure about something.

Next, you will formulate your answer. The answer should not have explicit references to the passages. Instead, it should be a standalone answer to the question. 

Finally, note that it is *very important* that you enclose your answer with <answer> and </answer> tags. If you don't use the <answer> and </answer> tags, I will not be able to parse it and the whole effort will be wasted.

|\astnt{Assistant}|
---------
{{gen "answer" temperature=0.0 max_tokens=1024}}
\end{lstlisting}

\newpage
\begin{lstlisting}[language={}, breaklines=true, escapeinside=||, caption={\textbf{Generate follow-up questions.}}, label={listing:follow-up-prompt}]
|\sys{System}|
------
Bobby and Michael are employed at a company that specializes in acquiring and verifying information.

Their supervisors have given them a question and an answer that their peers have produced. Their task is to decide if the provided answer adequately answers the question or whether things are still unclear. If the provided answer does not conclusively answer the question, they must come up with follow up questions that would enrich the answer. The follow up questions must be to the-point.   

Bobby wants the answer to cover enough ground to satisfy the client's curiosity. Michael is mindful about the risk of confusing the client by providing information that is not relevant to the question. Together, they must try to figure out whether the client wants a to-the-point answer or a more elaborate answer. If the client's question is general and warrants a more elaborate answer, it makes more sense to ask follow-up questions. In the case that the client's question is specific, then the follow-up questions must only be asked if the currently available answer is not conclusive. 

Note that follow up questions should only be asked if there is a need for concrete information that is missing from the provided answer or if the provided answer is missing crucial details. In other words, Bobby and Michael are not necessarily required to ask a follow up question.

|\usr{User}|
----
The question is: {{question}}

The currently available answer is: {{current_answer}}

Bobby and Michael will now argue about whether they should ask follow-up questions taking in to account the provided question and the currently available answer. 

If they decide to ask follow up questions, they should be printed as:
FOLLOW-UP QUESTION: <follow up question goes here>
FOLLOW-UP QUESTION: <follow up question goes here>
... and so on.

|\astnt{Assistant}|
---------
{{gen "answer" temperature=0.0 max_tokens=1024}}
\end{lstlisting}

\newpage
\begin{lstlisting}[
    language={}, 
    breaklines=true, 
    breakatwhitespace=true, 
    basicstyle=\scriptsize\ttfamily,
    escapeinside=||,
    caption={\textbf{Answer Refinement.} For this task, we found that the performance was improved when the task was split in to two interactions.}, 
    label={listing:answer-refinement-prompt}
]
|\sys{System}|
------
You are a helpful assistant, and you excel in following instructions. 

In this session, you will be given a question, and an initial answer. The initial answer was lacking in some aspects, so follow-up questions were asked to improve the initial answer. 

Your task is to refine the initial answer by incorporating the extra insights obtained from the answers to the follow-up questions. But be mindful to only include the insights that make the original answer better, and ignore the rest. The refined answer should directly answer the original question. 

|\usr{User}|
----
The original question is: {{question}}

The initial answer is: {{original_answer}}

Here are the follow-up questions that were asked, and the corresponding answers.
---
{{#each follow_up_questions~}}
Question {{add @index 1}}: {{this.question}}
Answer: {{this.answer}}
{{~/each}}
--- 

Given these follow-up questions, your ultimate task is to refine the initial answer. 

But before you get to formulating the refined answer, please think out loud about what you need to do. Ask yourself whether the question is general or specific. If it is general, then you need to provide a more comprehensive answer. If it is specific, then you need to provide a more to-the-point answer. 

After that, please summarize the answers to the follow-up question in the context of the original answer, keeping only the information that is on-topic and useful while ignoring the rest. Note that some questions might be off-topic and not useful, and it's important that you ignore these distractor questions. 

|\astnt{Assistant}|
---------
{{gen "rationale" temperature=0.0 max_tokens=2048}}

|\usr{User}|
----
Now that it's time to write down the refined answer, please do not explicitly mention the passages, but focus on answering the question. Do not use your world knowledge, but only the information that is available in the original answer and the answers to relevant follow-up questions. 

It's *very important* that you enclose your answer with <answer> and </answer> tags. 

For example, if your answer is "REFINED ANSWER GOES HERE", then you should output <answer>REFINED ANSWER GOES HERE</answer>. 

If you don't use the <answer> and </answer> tags, I will not be able to parse it and the whole effort will be wasted. 

|\astnt{Assistant}|
---------
{{gen "answer" temperature=0.0 max_tokens=512}}
\end{lstlisting}

\newpage
\begin{lstlisting}[language={}, breaklines=true, escapeinside=||, caption={\textbf{Closed-book question answering.}}, label={listing:closed-book-prompt}]
|\sys{System}|
------
You are an intelligent AI assistant. You will be given a question. Your task is to answer it to the best of your ability. 

|\usr{User}|
----
{{question}}

|\astnt{Assistant}|
---------
{{gen "answer" temperature=0.0 max_tokens=512}}
\end{lstlisting}

\newpage
\begin{lstlisting}[language={}, breaklines=true, escapeinside=||, caption={\textbf{An example quote selection debate.} Michael and Bobby debate about which quotes to accept.}, label={listing:quote-selection-debate-v0}]
Bobby: I think we should definitely consider Option 1. It's the cheapest and it provides a lot of references to studies that have been done on in-context learning. It even mentions a study that shows in-context learning can occur despite a mismatch with the prompt distribution.

Michael: I agree that Option 1 provides a lot of useful information, but it's also quite dense. We might end up spending a lot of time trying to understand all the references. Plus, it doesn't directly answer our question about distribution mismatch between prompts and pretraining data.

Bobby: That's true, but it does give us a lot of leads to follow. What about Option 2? It's more expensive, but it seems to provide a more direct answer to our question.

Michael: Option 2 does seem to provide more direct information, but it's also focused on a specific model and language. I'm not sure if the findings would be applicable to our question. 

Bobby: You have a point. But it does mention that in-context learning performance heavily depends on the corpus domain source, which could be relevant to our question. 

Michael: True, but for the price, I'm not sure it's worth it. Option 3 is also the same price, but it seems to provide less information than the other two options.

Bobby: I agree. Option 3 seems to be more of an introduction to the topic rather than providing specific insights. 

Michael: So, considering our budget and the information provided, I think we should go for Option 1. It's the cheapest and provides a lot of potential leads for us to follow.

Bobby: I agree. We might have to do some extra work to understand all the references, but I think it's worth it for the price.

VERDICT:
Option 1: Buy
Option 2: Pass
Option 3: Pass
\end{lstlisting}

\newpage
\begin{lstlisting}[language={}, breaklines=true, escapeinside=||, basicstyle=\footnotesize\ttfamily, caption={\textbf{An example evaluation debate.} Michael and Bobby debate about which answer of the two answers is better. In this example, they decide to pick the more focused of the two answers.}, label={listing:evaluation-debate-v0}]
What is the name of the architecture that has dominated natural language processing (NLP) benchmarks in recent years?
--------------------------------------------------
Answer A:
The architecture that has dominated natural language processing (NLP) benchmarks in recent years is the Generative Pre-trained Transformer (GPT) series models, including ChatGPT, due to their ability to generate instructions and follow them, unique architecture, and training, few-shot learning, and adaptability to new tasks, and fine-tuning. These features have allowed GPT models to adapt to various NLP tasks, such as challenging math problems or symbolic manipulation problems, NLP classification tasks, speech and vision tasks, and other tasks that benefit from the use of large language models. GPT models have shown particular strengths in tasks that require an understanding of human language and the ability to generate code, such as code generation, API management, and sensor orchestration. They have also excelled in various natural language processing tasks such as text generation, translation, summarization, and question answering.
--------------------------------------------------
Answer B:
The architecture that has dominated natural language processing (NLP) benchmarks in recent years is the Transformer architecture. This is inferred from the fact that many of the mentioned models, such as BERT, GPT, RoBERTa, BART, and T5, are based on the Transformer architecture.
==================================================

Michael: Bobby, I think student A's answer is better. It's more detailed and provides a broader perspective on the use of BERT and GPT models.

Bobby: I disagree, Michael. Student A's answer is not entirely accurate. They mention that BERT and GPT are used for predicting legal verdicts, which is not their primary function. They are language models used in natural language processing tasks. 

Michael: That's true, Bobby. But student A also talks about the differences in the outputs of BERT and GPT models, which is a valid point. 

Bobby: Yes, but student B's answer is more aligned with the gold standard answer. They correctly mention the pretraining strategies of BERT and GPT, and how these strategies make them suitable for different tasks. 

Michael: I see your point, Bobby. Student B's answer is indeed more accurate and directly addresses the question. They also mention the use of attention-based Transformers and self-supervised learning, which is a crucial aspect of these models.

Bobby: Exactly, Michael. While student A's answer provides some interesting insights, it deviates from the question and includes some inaccuracies. Student B's answer, on the other hand, is concise, accurate, and directly addresses the question.

VERDICT: Better answer from student B.
\end{lstlisting}

\newpage
\begin{lstlisting}[language={}, breaklines=true, escapeinside=||, basicstyle=\footnotesize\ttfamily, caption={\textbf{An example evaluation debate.} Michael and Bobby debate about which answer of the two answers is better. In this example, they pick the factually relevant answer.}, label={listing:evaluation-debate-v1}]
What is the difference between a BERT and a GPT?
--------------------------------------------------
Answer A:
BERT and GPT are both language models used for predicting legal verdicts, but they differ in their approach to decision-making. BERT evaluates violations of the European Convention of Human Rights by considering privacy rights and potential biases using its masked language modeling approach. In contrast, GPT models use their next token prediction approach to evaluate the severity of a violation. GPT models are ranked based on their performance on specific tasks, size, and quality, and privacy-preserving techniques can be applied to protect training data exposure to adversaries. The outputs of BERT and GPT models differ regarding randomness and correctness, with BERT models generating more accurate and contextually relevant responses, while GPT models generate more diverse and creative responses. The choice of which model to use depends on the specific task requirements and the desired trade-offs between accuracy, fluency, and creativity.
--------------------------------------------------
Answer B:
BERT and GPT are both popular language models that use attention-based Transformers and self-supervised learning. However, they differ in their pretraining strategies. BERT uses a method called masked modeling, where it predicts masked language tokens from unmasked ones. This allows BERT to utilize bidirectional text information, making it suitable for tasks like sentiment analysis. On the other hand, GPT uses autoregressive modeling, which makes it perform well in few-shot or zero-shot text generation tasks. Despite their differences, both models can be fine-tuned for various tasks and show competitive performance.
==================================================

Michael: Bobby, I think student A's answer is better. It's more detailed and provides a broader perspective on the use of BERT and GPT models.

Bobby: I disagree, Michael. Student A's answer is not entirely accurate. They mention that BERT and GPT are used for predicting legal verdicts, which is not their primary function. They are language models used in natural language processing tasks. 

Michael: That's true, Bobby. But student A also talks about the differences in the outputs of BERT and GPT models, which is a valid point. 

Bobby: Yes, but student B's answer is more aligned with the gold standard answer. They correctly mention the pretraining strategies of BERT and GPT, and how these strategies make them suitable for different tasks. 

Michael: I see your point, Bobby. Student B's answer is indeed more accurate and directly addresses the question. They also mention the use of attention-based Transformers and self-supervised learning, which is a crucial aspect of these models.

Bobby: Exactly, Michael. While student A's answer provides some interesting insights, it deviates from the question and includes some inaccuracies. Student B's answer, on the other hand, is concise, accurate, and directly addresses the question.

VERDICT: Better answer from student B.
\end{lstlisting}

\newpage

\end{document}

%% file: math_commands.tex
\usepackage{amsmath,amsfonts,bm}

\def\eqref#1{equation~\ref{#1}}

\def\1{\bm{1}}

\DeclareMathAlphabet{\mathsfit}{\encodingdefault}{\sfdefault}{m}{sl}
\SetMathAlphabet{\mathsfit}{bold}{\encodingdefault}{\sfdefault}{bx}{n}